\documentclass[a4paper]{article}

\usepackage[english]{babel}
\usepackage[utf8]{inputenc}
\usepackage{amsmath}
\usepackage{graphicx}
\usepackage[colorinlistoftodos]{todonotes}
\usepackage{listings}
\title{Template Matching based Object Detection Using HOG Feature Pyramid}

\author{Anish Acharya\\Dept. of Electrical Engineering and Computer Science\\ University of California-Irvine\\anisha@uci.edu}

\begin{document}
\maketitle
\section{Introduction}
The Object detection based on histogram of oriented gradients was introduced in the context of detecting humans in images [1]. Detecting human subject in the image is a difficult task due to the fact that they appear in different appearance and poses. Thus the need to come up with a feature that allows human dorm to be detected cleanly even in cluttered background under different illumination. So, it deals with some of the key issues in object detection.
\begin{itemize}
\item varied poses
\item different backgrounds
\item different appearance and clothing
\item scale variance
\end{itemize}

\section{Histogram of Oriented Gradients}
The main idea conveyed by Dalal et.al.[1]the method of feature extraction is based on evaluating well normalized local image gradient orientation in a dense grid over the image.\\This is the feature extraction step and the novelty of this approach[1].
Let us go through the sttep by step procedure to implement this.
Let us first take an example image, which is shown in FIg.1
\begin{figure}
\centering
\includegraphics[width=1\textwidth]{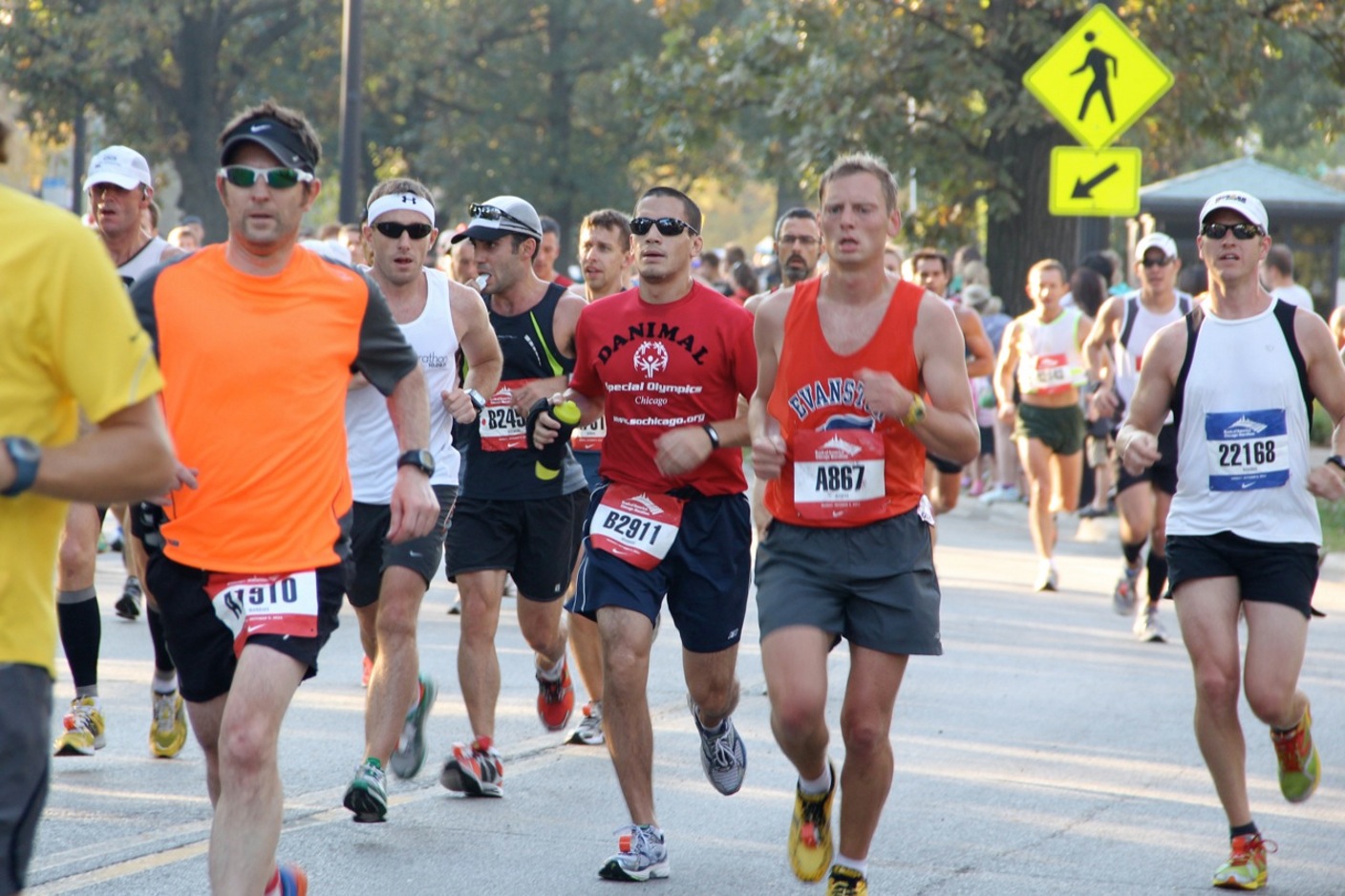}
\caption{\label{fig:frog}The initial image}
\end{figure}
\begin{enumerate}
\item Compute the gradient magnitude and orientation:\\ The function named mygradient.m does this computation. This is included in the appendix.
There are several ways to compute the gradient. Here, I chose to use the simplest centered 1-D gradient operator with no gaussian smoothing as reported in [1]. The functions are enlisted in the appendix. Here we just put in the series of commands so that it gives a good understanding of the operational method. 
\item Find the HOG features:\\ So, here we divide the entire image into 8*8 blocks and we decide to find the gradient orientations under 9 bins that means we chose 9 directions. So, basically we selected the possible orientations to be in these nine angular blocks -pi:2*pi/9:pi. 
\\ Then the idea was to calculate how many edges with orientation k where $k\in 1:9 $ belong to each of those 8*8 blocks. We also add the constraint that in order to qualify as an edge the response has to cross a pre-defined threshold which in this case we chose to be 10\% of the max of magnitude response of the edge detector. 
\\ hog.m function does all thtese computations mentioned and comes up with the HOG features. hog.m is also included in the appendix. hogdraw.m function helps us visualize the HOG view. Fig.2 showws the visualization of HOG features i.e. the HOGgles view. [2]
\begin{lstlisting}
clear all;
I=imread('test0.jpg');
figure, imagesc(I)
I=rgb2gray(I);
I=im2double(I);
ohist=hog(I);
V=hogdraw(ohist,15);
figure, imagesc(V);
\end{lstlisting}
\begin{figure}
\centering
\includegraphics[width=1.25\textwidth]{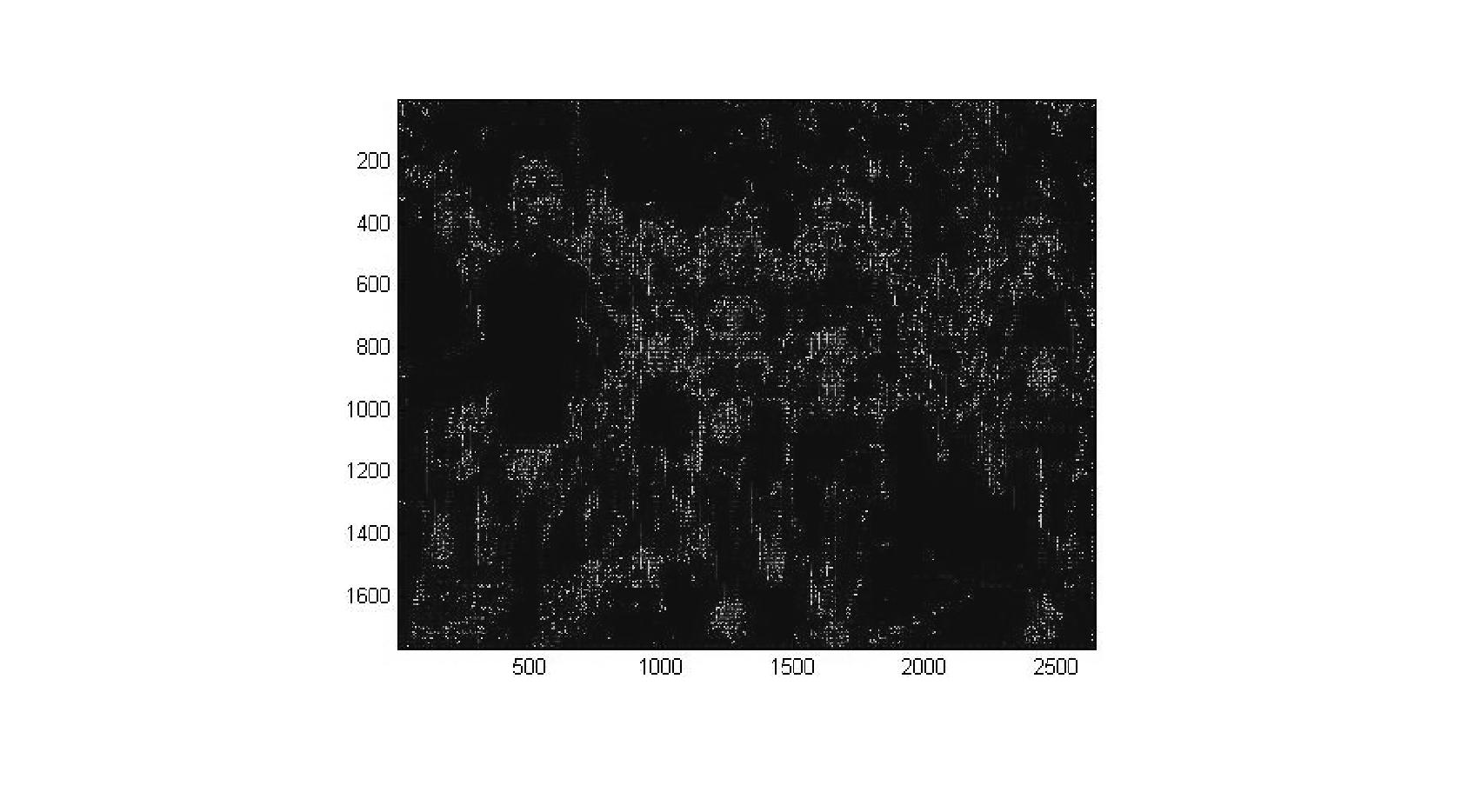}
\caption{\label{fig:frog}HOG visualization of Fig.1}
\end{figure}
\item Detection:\\Here we are using a Tempalte matching based detection where we choose a patch from the test image and use cross-correlation to find the places in the test image where we get a strong response and declare those places as matched response i.e. we claim to detect the given template at these location of the test image. [3] Here we output our most confident top n number of detections where we can select n beforehand. Let us see what happens if we do it using Fig.1 as test image and the selected patch is shown in Fig.3. So, we train our model with this patch in Fig. 1 and we now want to detect similar object from Fig. 4. The detection result is shown in Fig. 5. Here we enlist our top 5 detection where green is the most confident detection and confidence fades to red being the least confident detection among the 5 detections.
\begin{figure}
\centering
\includegraphics[width=1\textwidth]{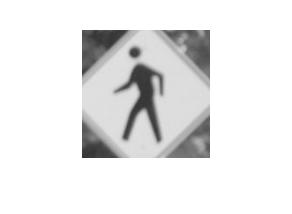}
\caption{\label{fig:frog}Template}
\end{figure}
\begin{figure}
\centering
\includegraphics[width=1\textwidth]{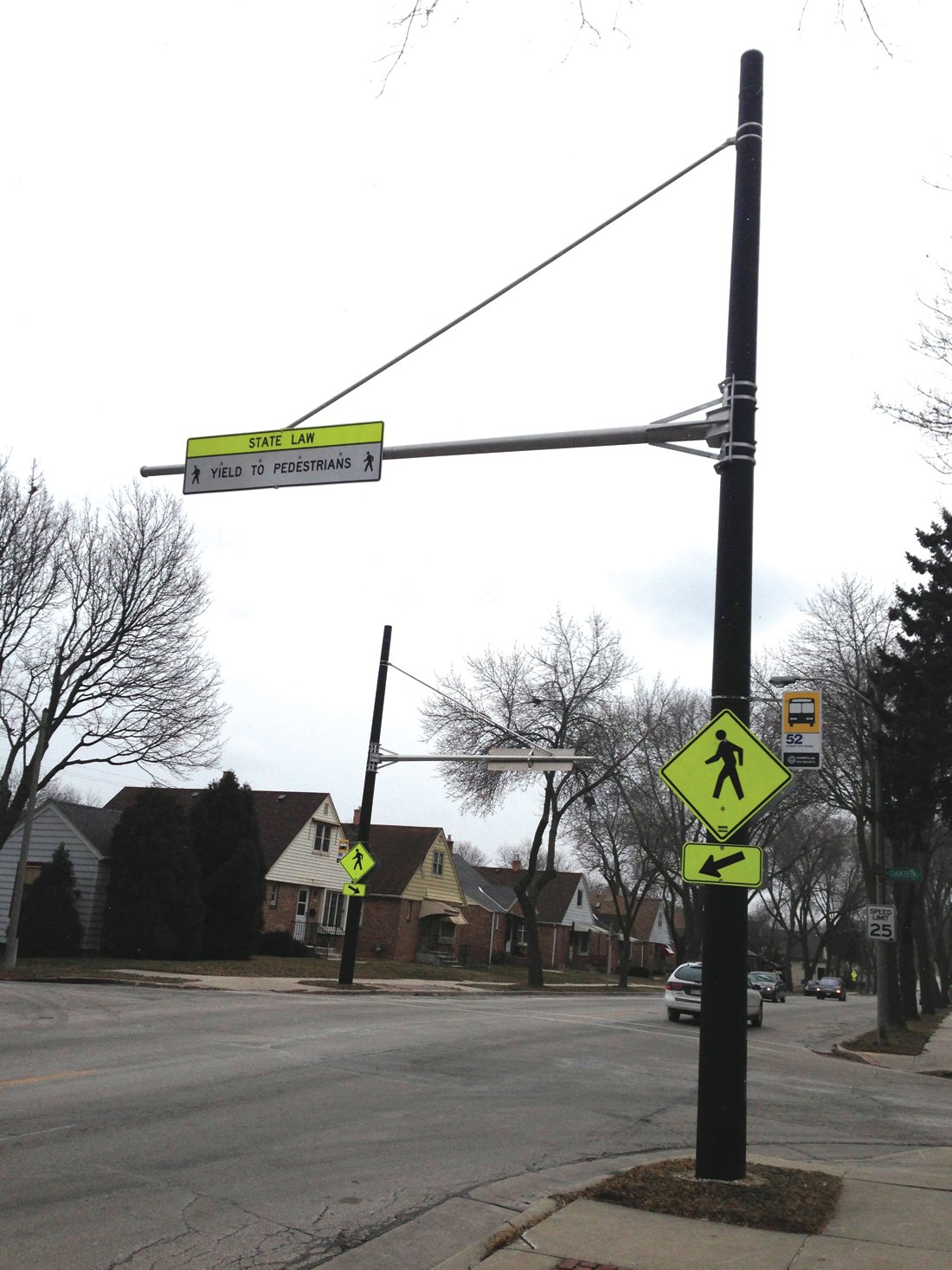}
\caption{\label{fig:frog}Test Image}
\end{figure}
\begin{figure}
\centering
\includegraphics[width=1\textwidth]{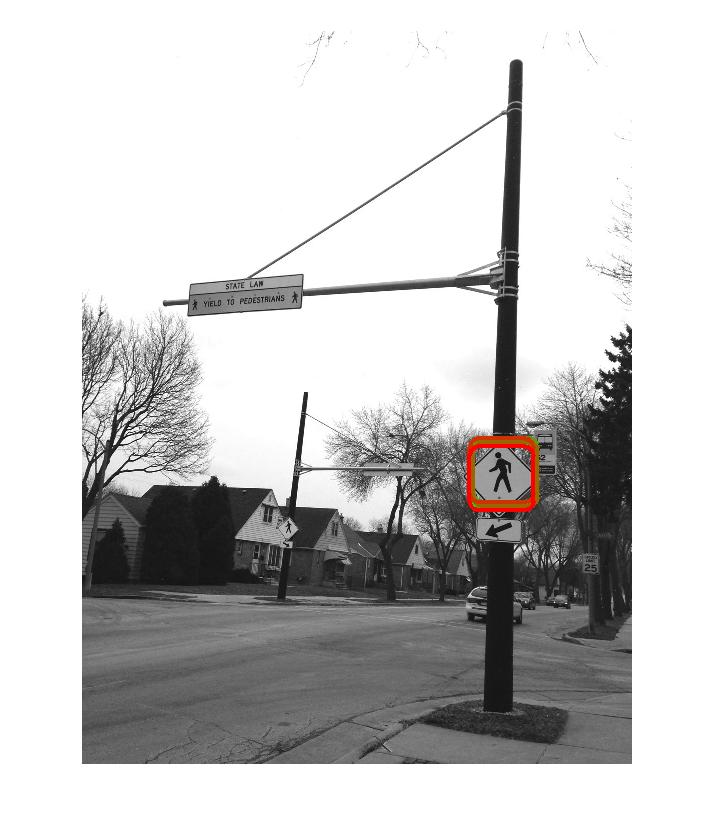}
\caption{\label{fig:frog}Detection}
\end{figure}
\item Discard overlapping detections\\
Now, as we notice from Fig.5 that the detector spits out top 5 detections but all are pretty overlapping as expected since the close by pixels of the best detection are also susciptible to high triggering of the cross correlation output. Thus, if we do not make any attempt to move around this problem we won't be able to detect multiple instances of the object rather will always output near our most confident guess.
\\So, we impose a constraint on this detection by thresholding the detections by eucledean distance i.e. if the detections are overlapping we won't include that detection. This is done through the isequal.m function which sets a flag if the next detection is not suffficiently distinct from the previous one.The resulted output is shown in Fig.6.
\begin{figure}
\centering
\includegraphics[width=1\textwidth]{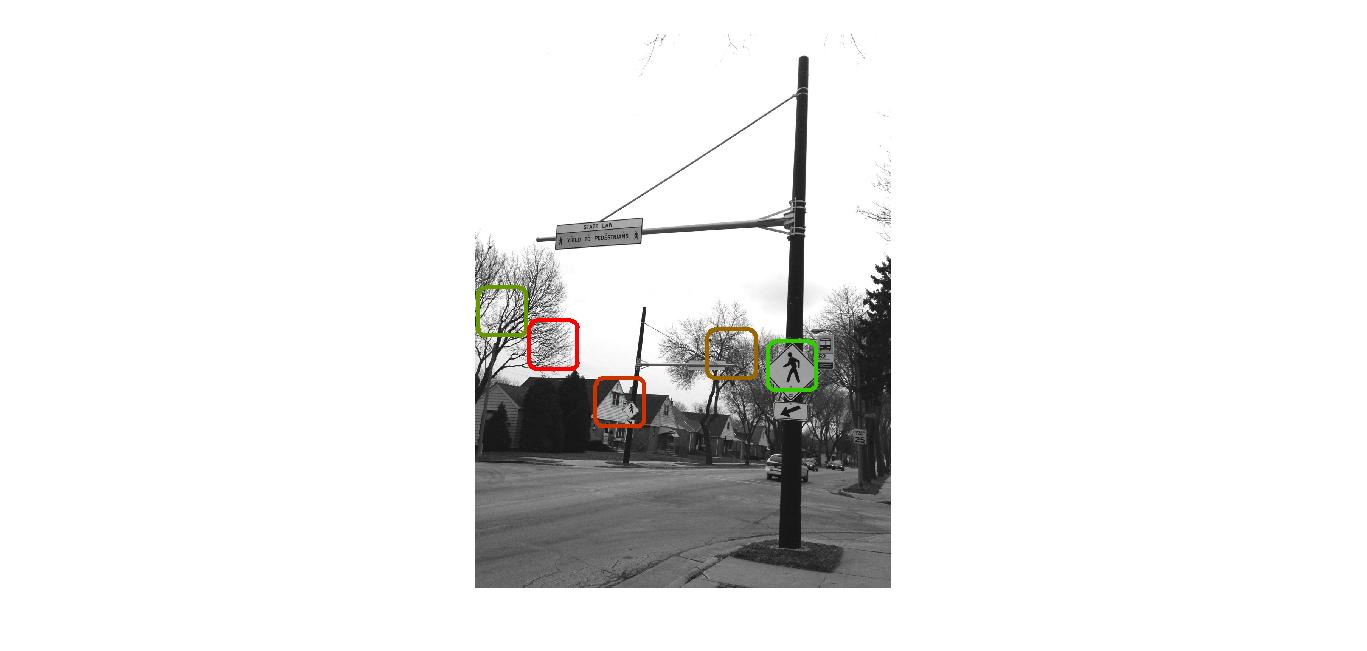}
\centering
\caption{\label{fig:frog}Detection after excluding overlapping detections}
\end{figure}
Here sufficiently close is a design choice. Now, since we have chosen our detection rectangle to be 64*64 we are setting the flag=1 if either of the two co-ordinates of the new detections is not 64*2=128 pixels apart from all the previously detected results. Now, as we see from Fig.6 that though our best guess i.e. the green box detects the best match but our less confident guesses are unable to detect similar objects at different scales as we would have liked it to. So, we should really movetowards the goal of scale invariant detection next.
Before, moving forward to scale invariance let us see the performance on some other test images containing the target object as shown in Fig7-9.  
\begin{figure}
\centering
\includegraphics[width=1\textwidth]{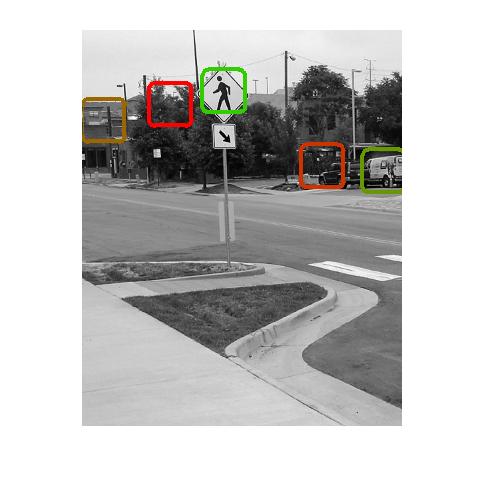}
\centering
\caption{\label{fig:frog}Detection after excluding overlapping detections}
\end{figure}
\begin{figure}
\centering
\includegraphics[width=1\textwidth]{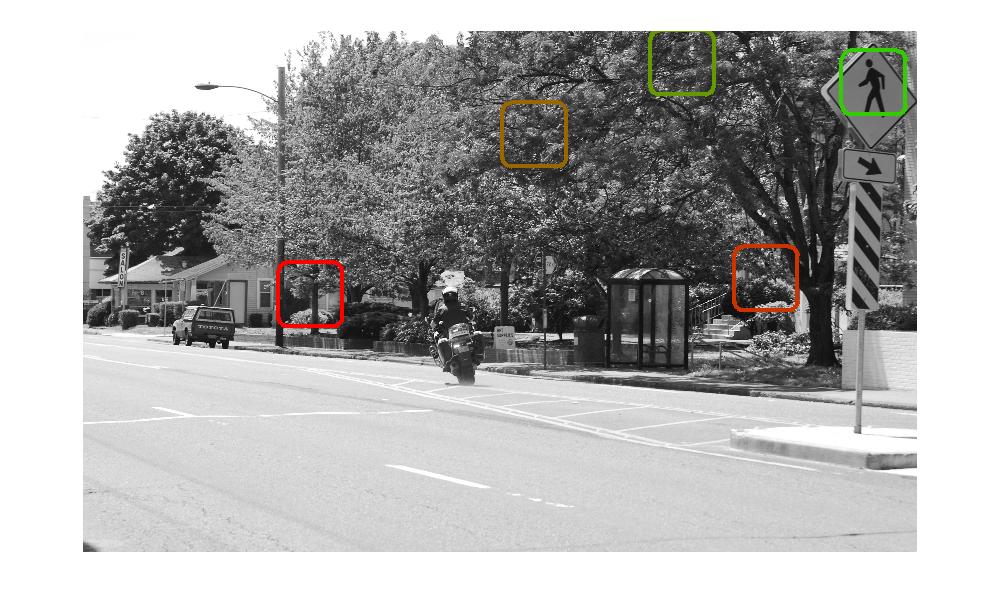}
\centering
\caption{\label{fig:frog}Detection after excluding overlapping detections}
\end{figure}
\begin{figure}
\centering
\includegraphics[width=1\textwidth]{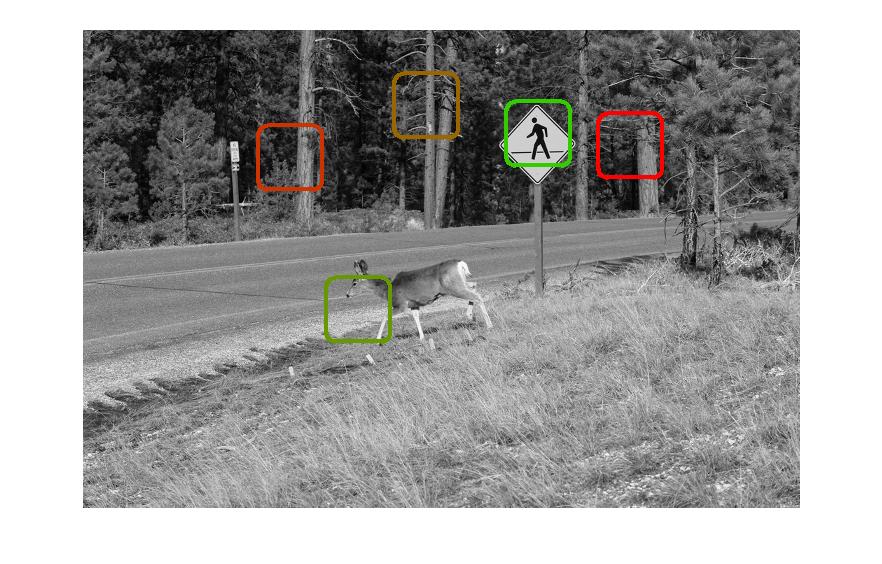}
\centering
\caption{\label{fig:frog}Detection after excluding overlapping detections}
\end{figure}
\item More Training Data:\\
Inspite of having one training image let us see what happens when we use multiple training points. So, as a first step let us generate 10 training templates from the same training image Fig.1 where avg. of the templates used for training are shown in Fig. 10 and let us see the test result in an test image shown in Fig. 11. which is much better than what we get as shown in Fig. 13 if trained with single training template shown in Fig. 12  
\begin{figure}
\centering
\includegraphics[width=1\textwidth]{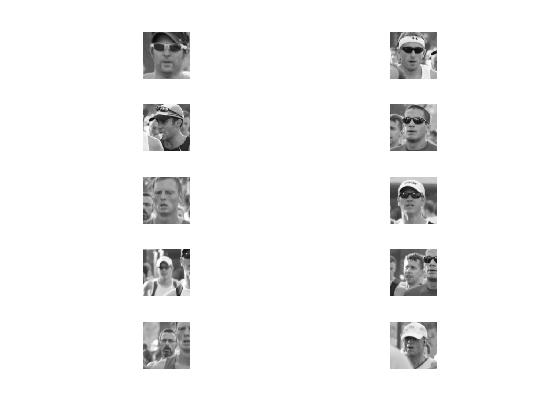}
\centering
\caption{\label{fig:frog}Training Templates extracted from Fig.1}
\end{figure}
\begin{figure}
\centering
\includegraphics[width=1\textwidth]{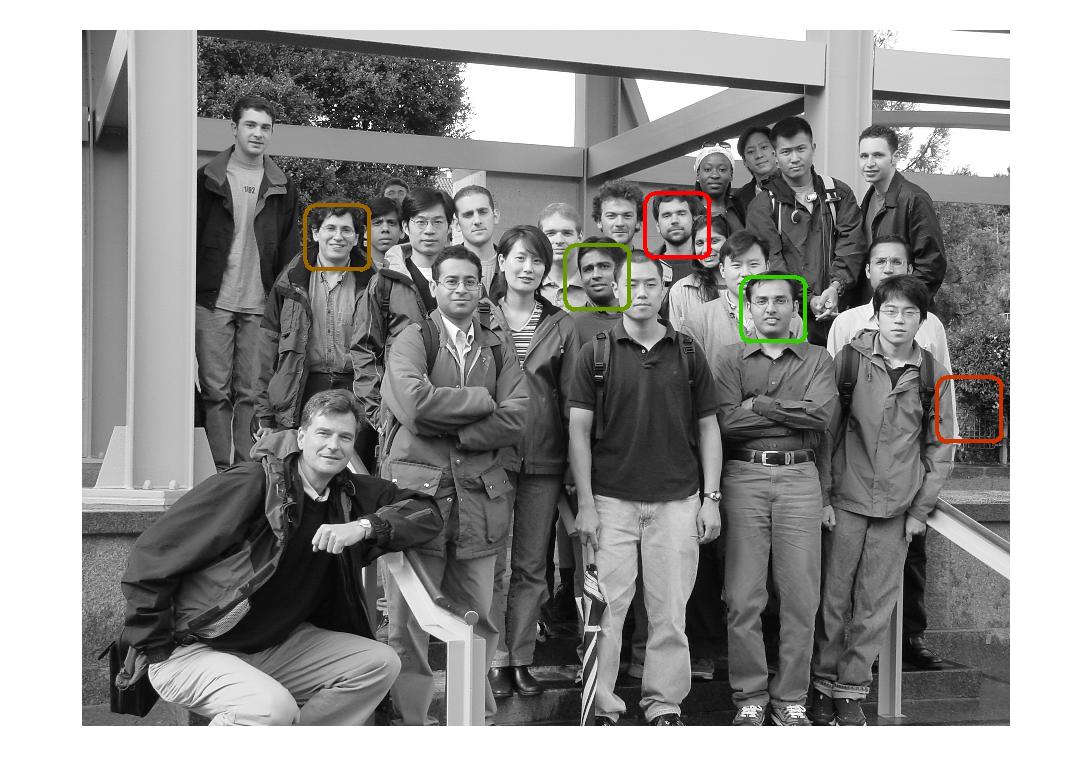}
\centering
\caption{\label{fig:frog}Detection results using Templates in Fig.10}
\end{figure}
\begin{figure}
\centering
\includegraphics[width=1\textwidth]{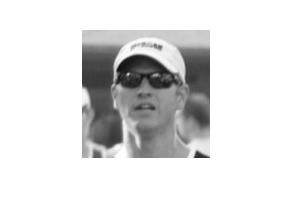}
\centering
\caption{\label{fig:frog}Template used to learn}
\end{figure}
\begin{figure}
\centering
\includegraphics[width=1\textwidth]{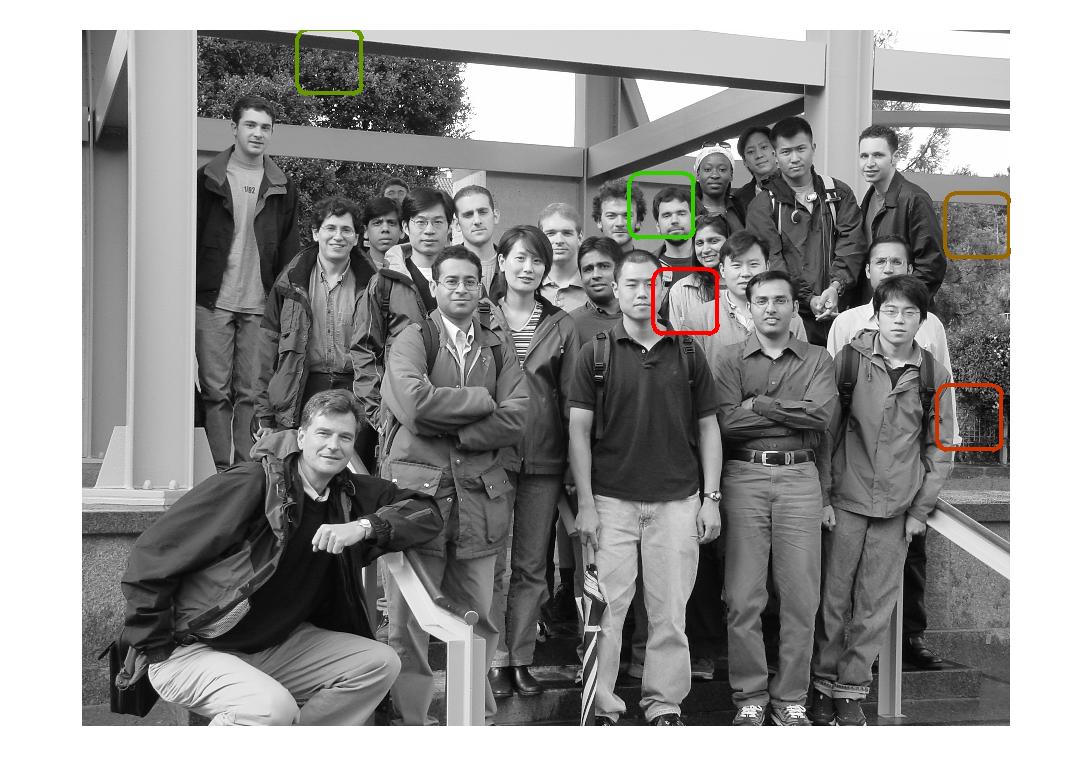}
\centering
\caption{\label{fig:frog}Detection results using Templates in Fig.12}
\end{figure}
\item What about using different Images to generate Training Set:\\Now, the idea is if we use a single image for generating multiple training points which is just an avg. of all these templates learnt from user clicks what we can do is we can use different images to learn different training points and it will intuitively have much less bias and should be a more general detector if we can train using sufficiently different images where we have the object in different pose and orientations and with varied outfit, background etc. After all that was the whole motivation behind this HOG based method. 
\\ However, one has to be very careful here. Since, we have the objects in different sizes and shape we might not want to take a fixed window of template rather we would like to draw a window around the training template instead of user click we want the user to choose a rectangle around the object. The same has been encoded using the detect\_script\_rect.m, improved\_detect\_script.m,my\_trainin.m,my\_trainin\_box.m,my\_trainin\_box2.m scripts attached in the appendix.
\\SO, now we use 5 Test Images shown in Fig. 14-18. 

\begin{figure}
\centering
\includegraphics[width=1\textwidth]{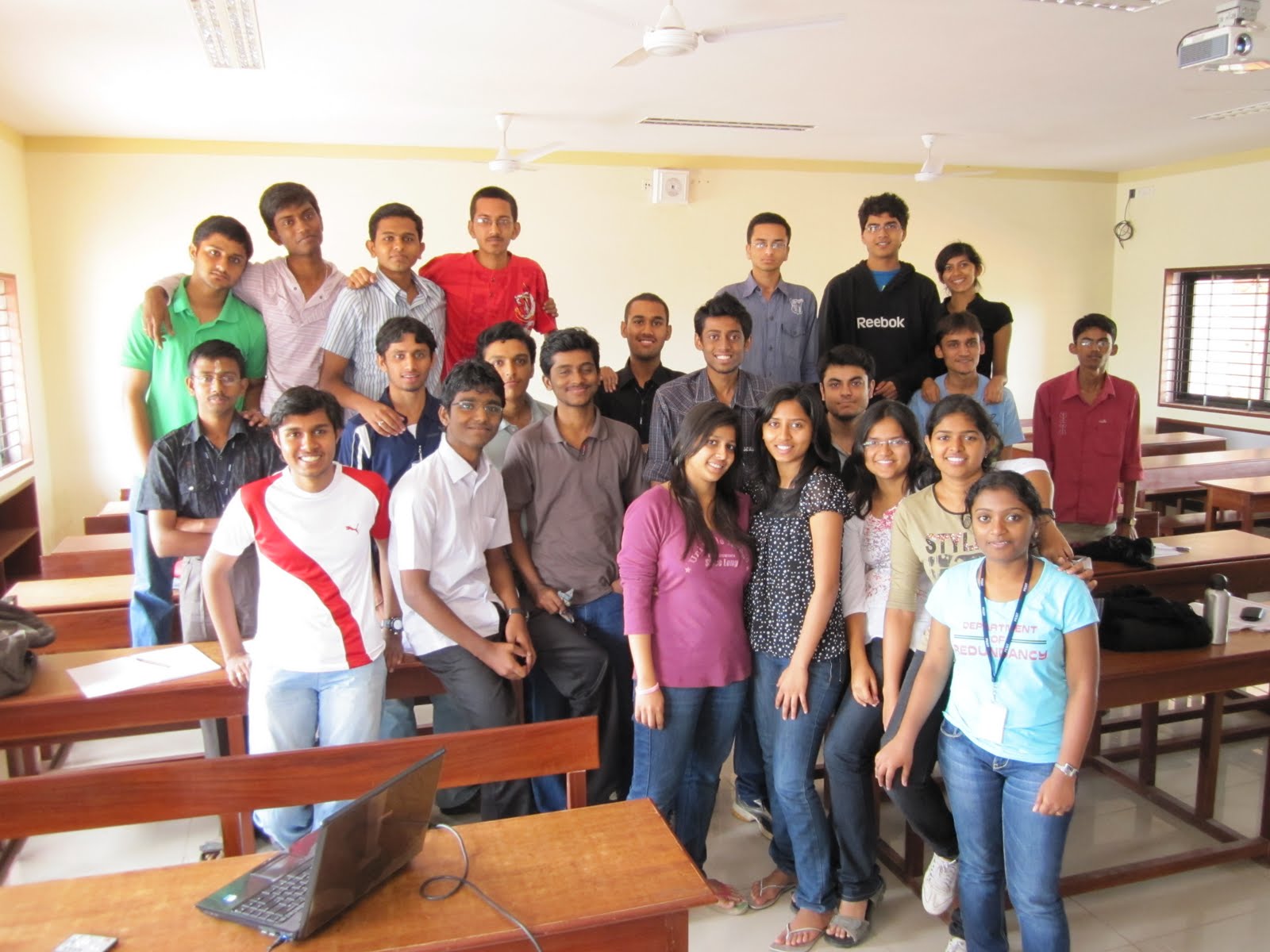}
\centering
\caption{\label{fig:frog}Training 1}
\end{figure}
\begin{figure}
\centering
\includegraphics[width=1\textwidth]{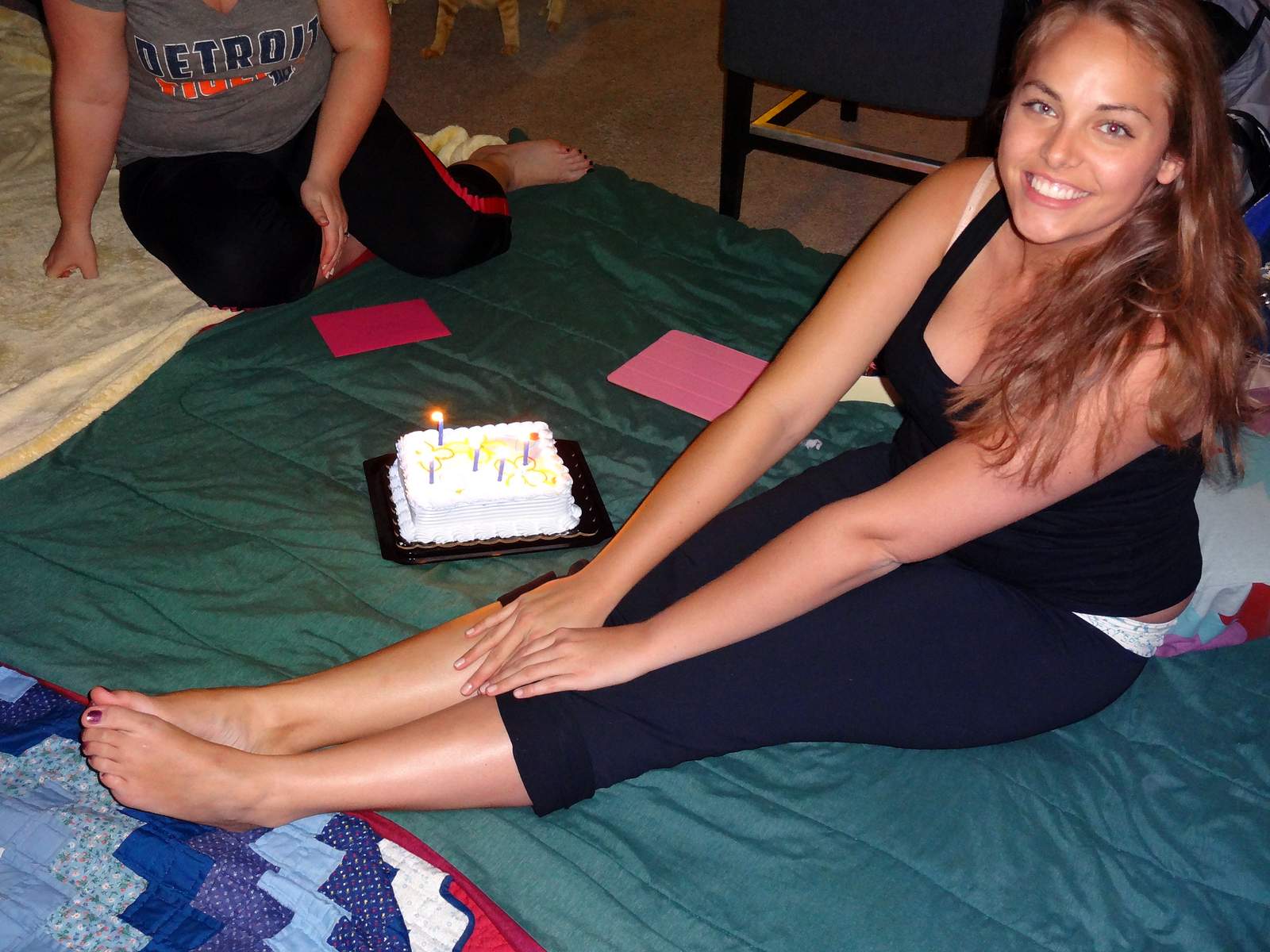}
\centering
\caption{\label{fig:frog}Training 2}
\end{figure}
\begin{figure}
\centering
\includegraphics[width=1\textwidth]{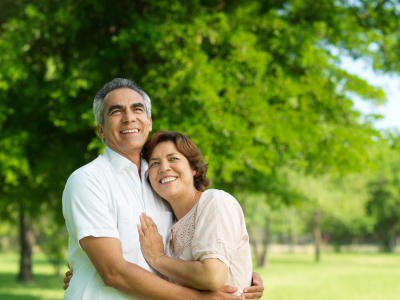}
\centering
\caption{\label{fig:frog}Training 3}
\end{figure}
\begin{figure}
\centering
\includegraphics[width=1\textwidth]{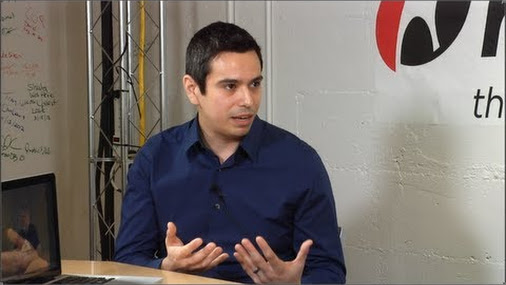}
\centering
\caption{\label{fig:frog}Training 4}
\end{figure}
\begin{figure}
\centering
\includegraphics[width=1\textwidth]{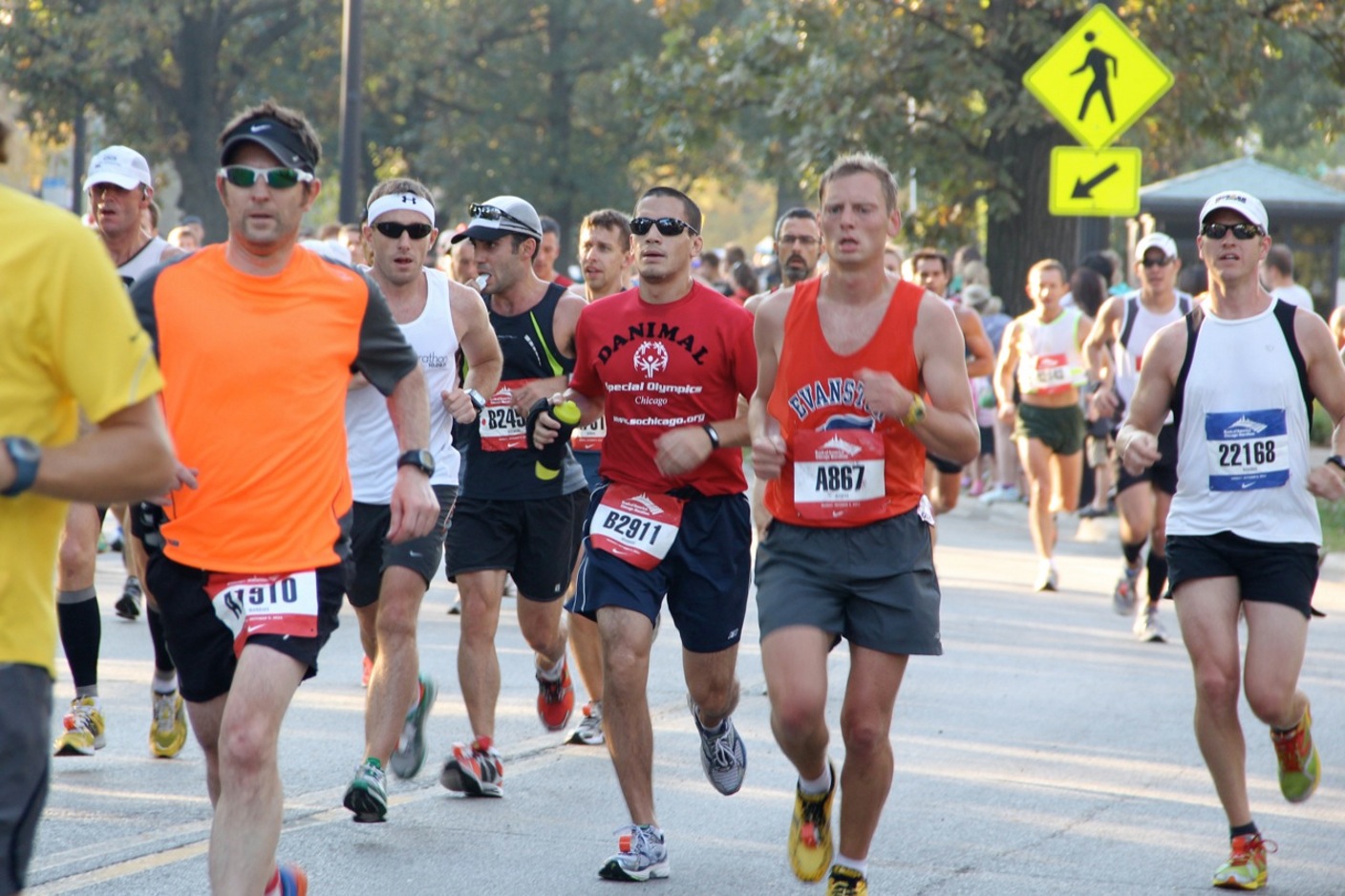}
\centering
\caption{\label{fig:frog}Training 5}
\end{figure}

Now, Fig. 19 shows the selected image patches drawn by the user from these examples.
\begin{figure}
\centering
\includegraphics[width=1\textwidth]{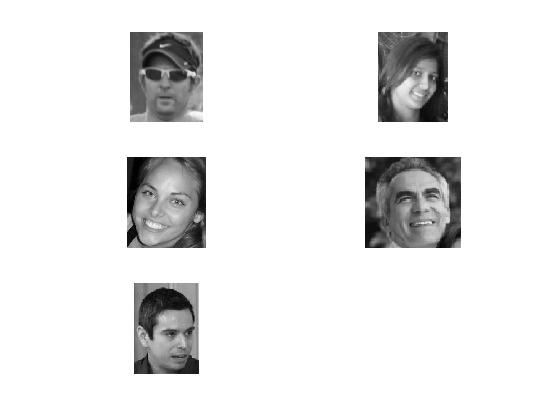}
\centering
\caption{\label{fig:frog}Training Templates}
\end{figure}
Now, using these templates let us see how we do over the Test image as shown in Fig. 20. This is done taking one sample from each image thus keeping Training Template=5. Now, we incorporate the idea of multiple clicks employed previously to choose multiple rectangular patches from the image thus now we take around 15 patches in total and train on these 15 training data and we obtain detection result as seen in Fig. 21. However, still we see that it is still detecting some similar objects from the background which are not necessarily the fces but since give high response at cross correlation thus a few misdetection. Thus, one intuition might be to tell the detector that if these are your ositive examples i.e. you are rewarded if you find these but, these are the wrong answers and you wil be penalized if you detect these. Thus, the basic idea is to generate a set of negative training examples. This can be anything at random. But, it should give better result if we can include a rich negative data set. Thus, we chose around 20 patches from each training images thus giving 100 negative training points and now we stick with what we started with i.e. one patch per image. Thus, we have a training set of 5positive and 100 negative examples. The detection result seems near perfect. It is shown in Fig. 22
\begin{figure}
\centering
\includegraphics[width=1\textwidth]{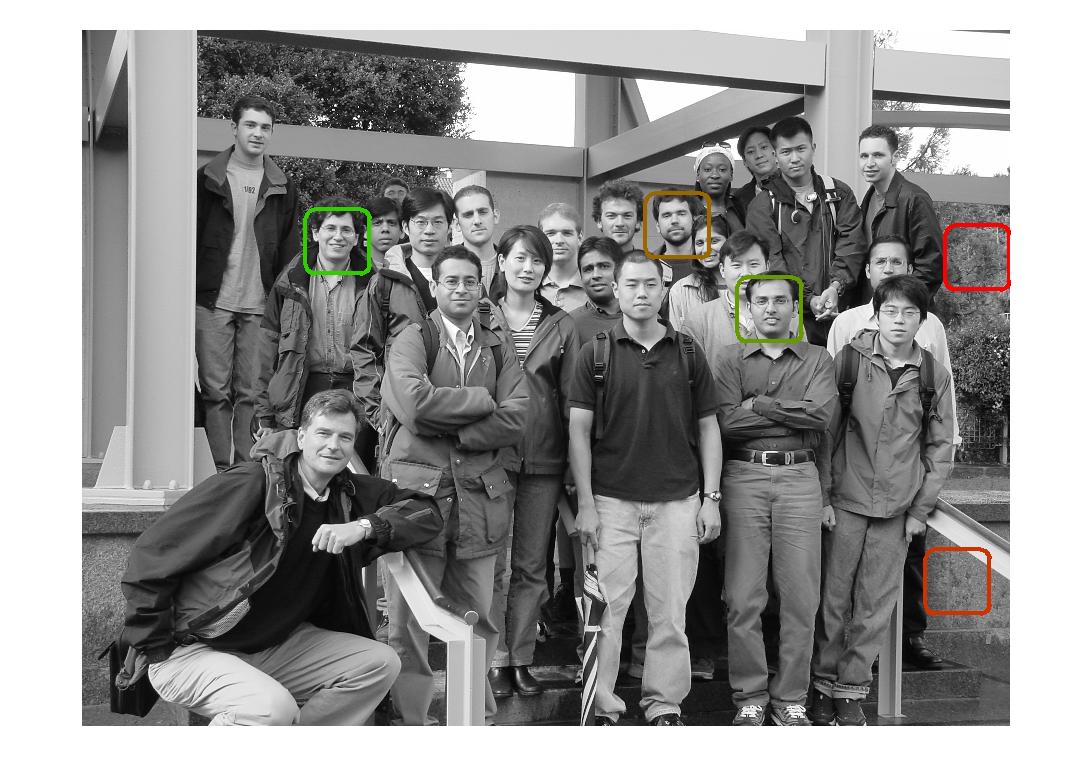}
\centering
\caption{\label{fig:frog}5 Positive Training Data}
\end{figure}
\begin{figure}
\centering
\includegraphics[width=1\textwidth]{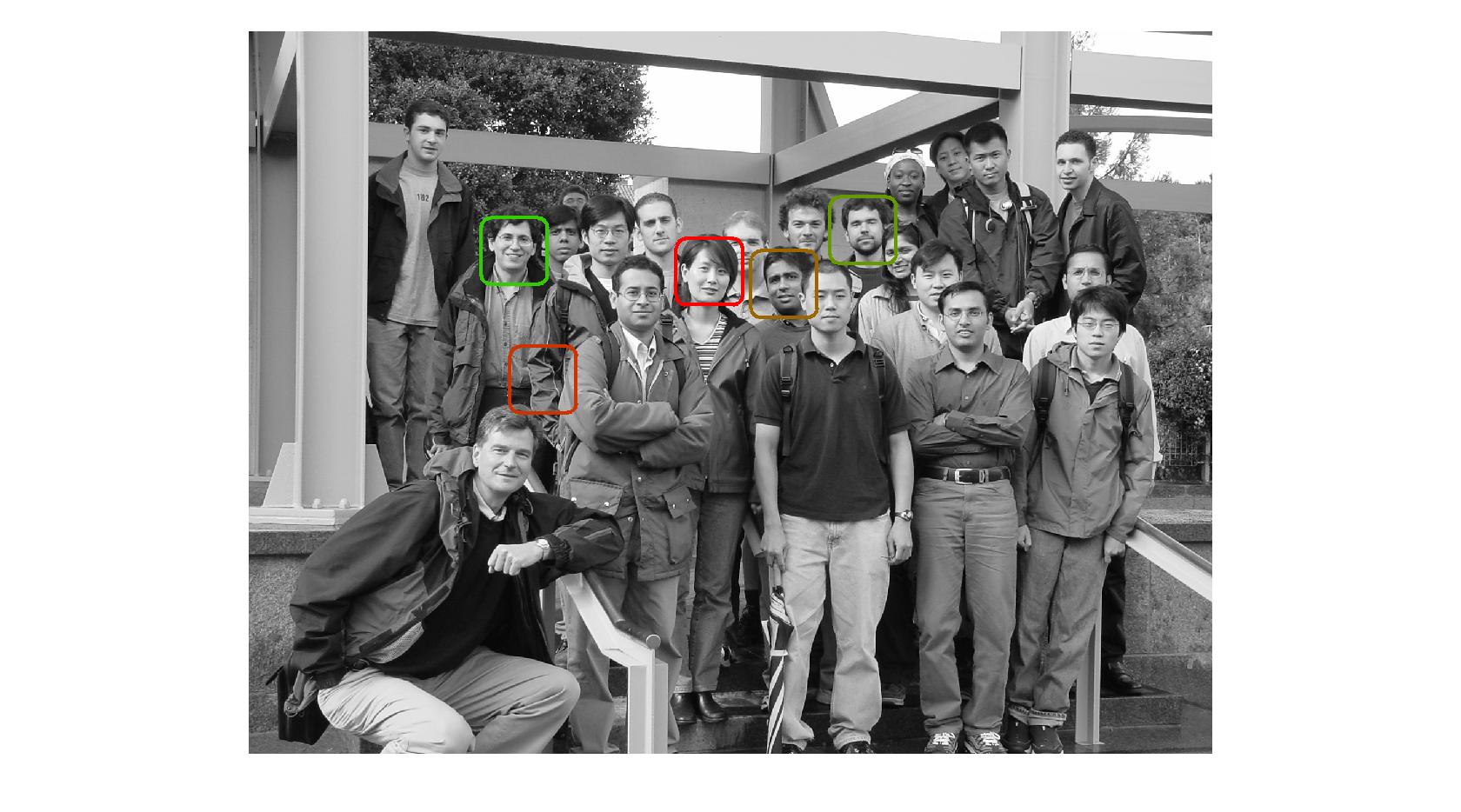}
\centering
\caption{\label{fig:frog}15 Positive Training Data}
\end{figure}
\begin{figure}
\centering
\includegraphics[width=1\textwidth]{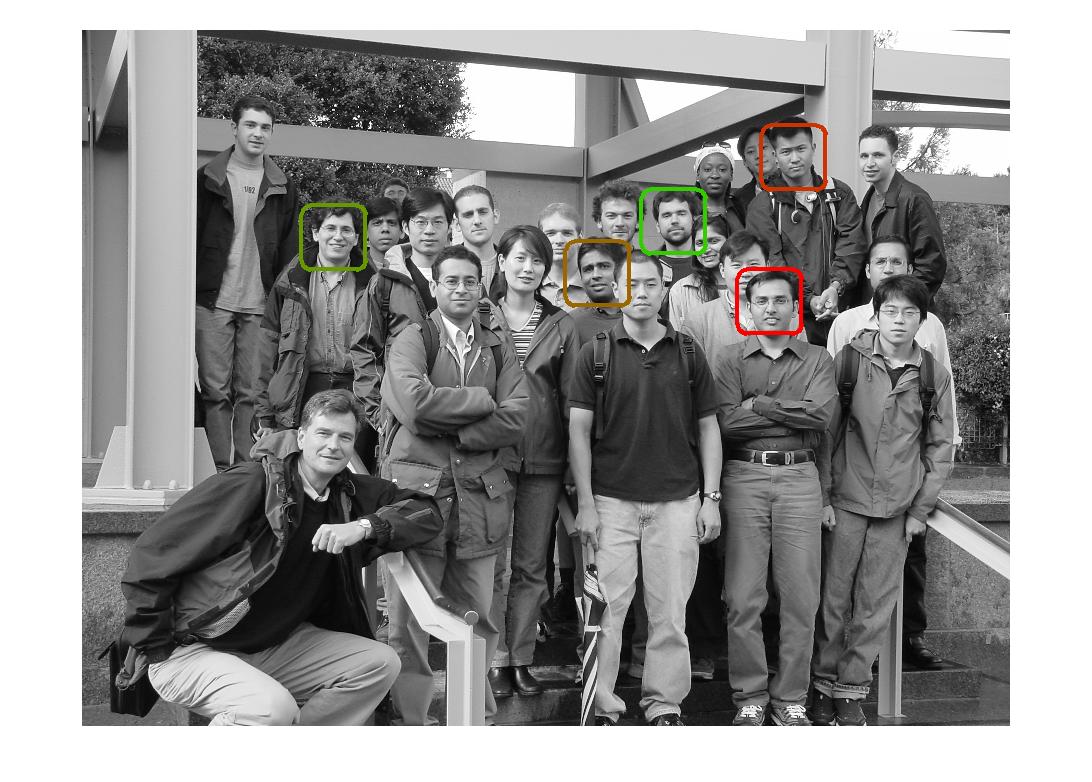}
\centering
\caption{\label{fig:frog}5 positive Training Data and 100 negative training data}
\end{figure}
We, now treat the test image as training image and the image test0.jpg i.e. Fig.1 as the test data and do the same operations all over again and see how does it perform on this new test image. 
So, we again do it with 1 positive training, 5 positive training and 5 positive + 100 negative training set results are shown in Fig. 23-25. 
\begin{figure}
\centering
\includegraphics[width=1\textwidth]{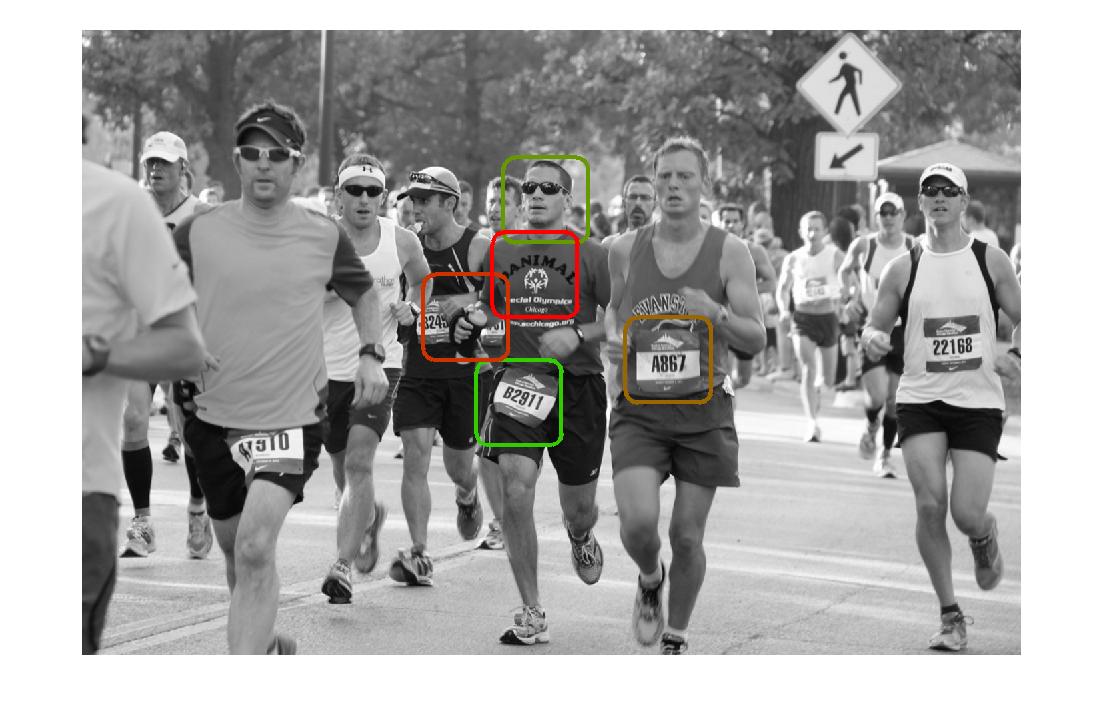}
\centering
\caption{\label{fig:frog}1 Positive Training Data}
\end{figure}
\begin{figure}
\centering
\includegraphics[width=1\textwidth]{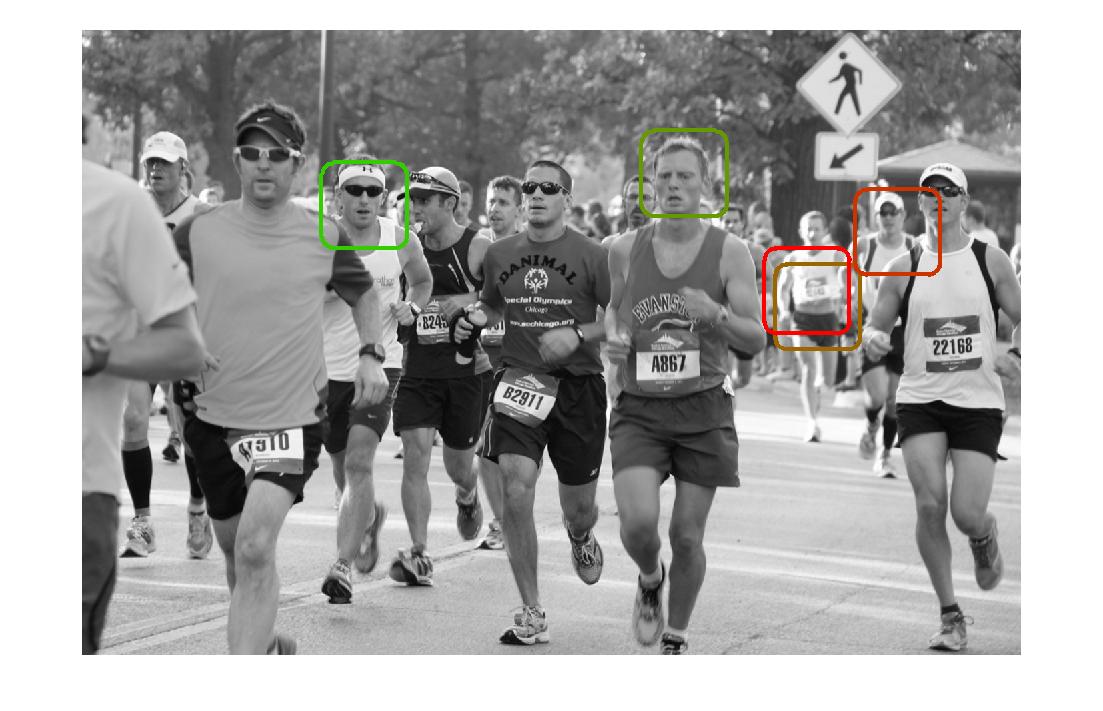}
\centering
\caption{\label{fig:frog}5 Positive Training Data}
\end{figure}
\begin{figure}
\centering
\includegraphics[width=1\textwidth]{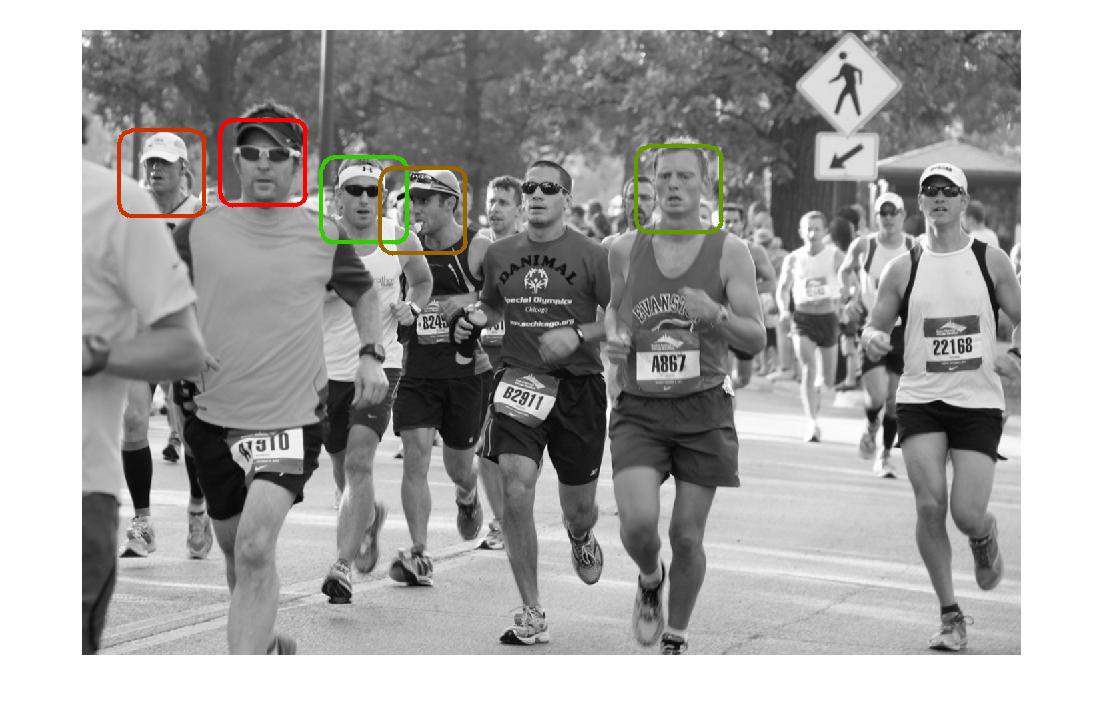}
\centering
\caption{\label{fig:frog}5 Positive and 100 negative Training Data}
\end{figure}

\item What About searching over different scales? \\
Now the idea is that suppose we want to find an object from different scenes. However, the major concern while using the template based matching is that the object might appear i different scales under different scenes. So, what we should really incorporate is to extend the idea to multi-scale detection. Now, the first thing we have to consider is that we learn only a single template at a particularly reasonable scale and then use that template to find objects at different scale. So, basically we are not changing the size of the template but we are constantly searching over different scales. The basic idea is to form a Image pyramid and thus a HOG pyramid as consequence. \\
So, we have the HOG feature space at different scales called {\bf The HOG feature pyramid} [4] and thus computing cross-correlation at all these different levels of the pyramid. \\
So, the idea is if we have objects at these different scales we can find these if we do detection at these different scales and then combine the results.
\\ Now, basically at each level of HOG pyramid we score the detections and find the locations as well then we rescale everything back to initial scale and spit out top scoring detections i.e. those top detections with highest score over all the levels of HOG pyramid. 
\\Let us start demonstrating it with a simple example where we have different scales of the same object and we choose the smallest as the template. Here, our assumption is we are not considering to detect objects smaller than a certain size. Thus, we keep on building the pyramid at a scale 0.5 untill the scaled image becomes smaller than the template. 
The result is shown in Fig. 26
Here, we chose the wheel as the object. 
\\ Here we did this using patch from the same object which we used to test so, it does not qualify to be a good result. So, we incorporate this multiscale idea with our previously built Object detector where we used 5 positive training points and 100 negative training points. So, basically we get the template averaging over the positive templates and averaging over the negative templates and taking their difference to be our template and matching it with the test i8mage over different scales and thus obtaining  top 5 detections based on the detection scores accumulated over different scales. The results are shown in Fig. 27 and Fig. 28. Though we do not improve any further as in the previous case also we detected all 5 correctly while using no scale invariance. Howver, we notice that in this case we come up with some faces at different scales which did not appear in the previous case. Thus, it establishes the fact that using scale invariance we can detect objects at different scales. So, when working in a much general case we are expected to get improved performance using this concept. 
\begin{figure}
\centering
\includegraphics[width=1\textwidth]{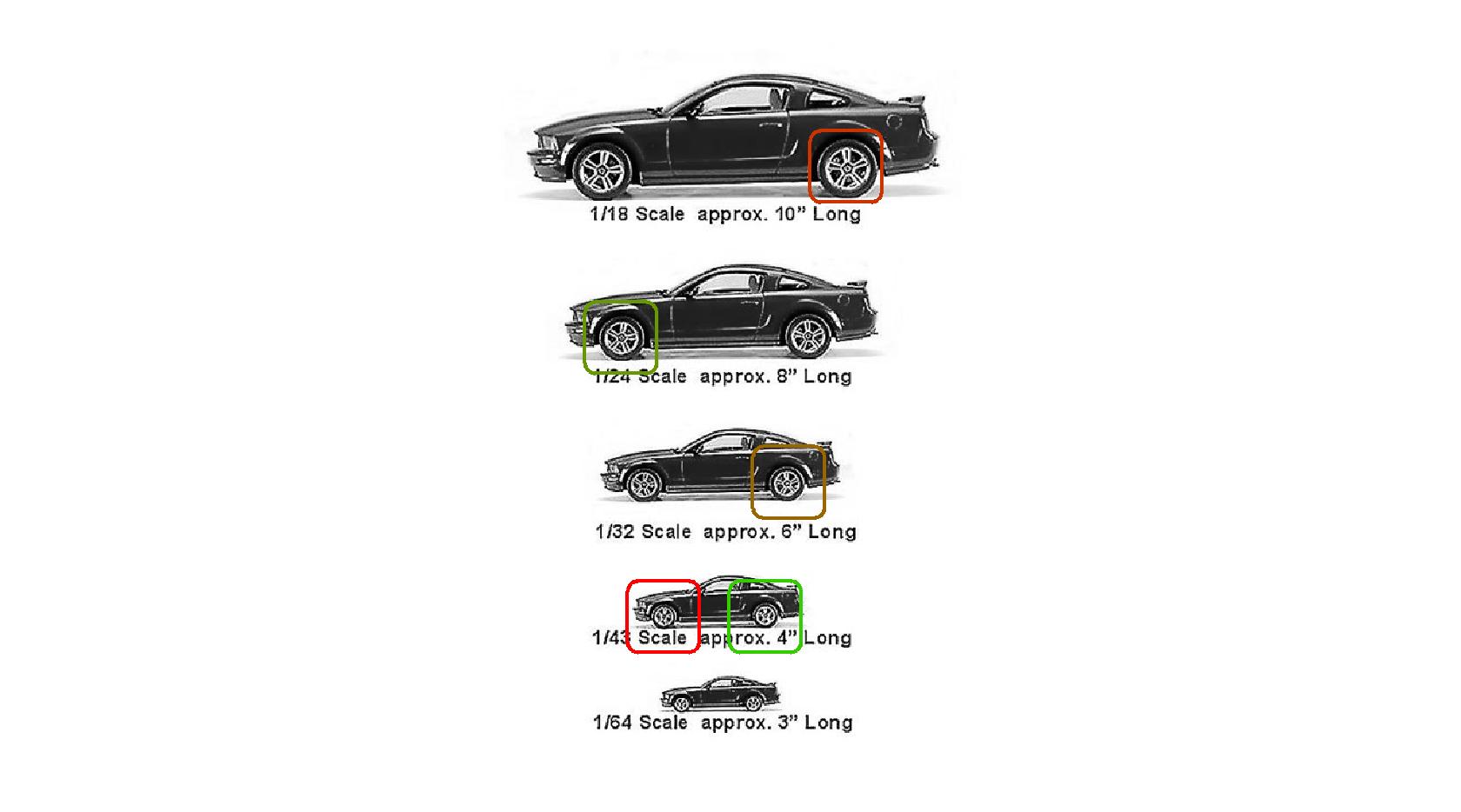}
\centering
\caption{\label{fig:frog}Detection at different scales}
\end{figure}
\begin{figure}
\centering
\includegraphics[width=1\textwidth]{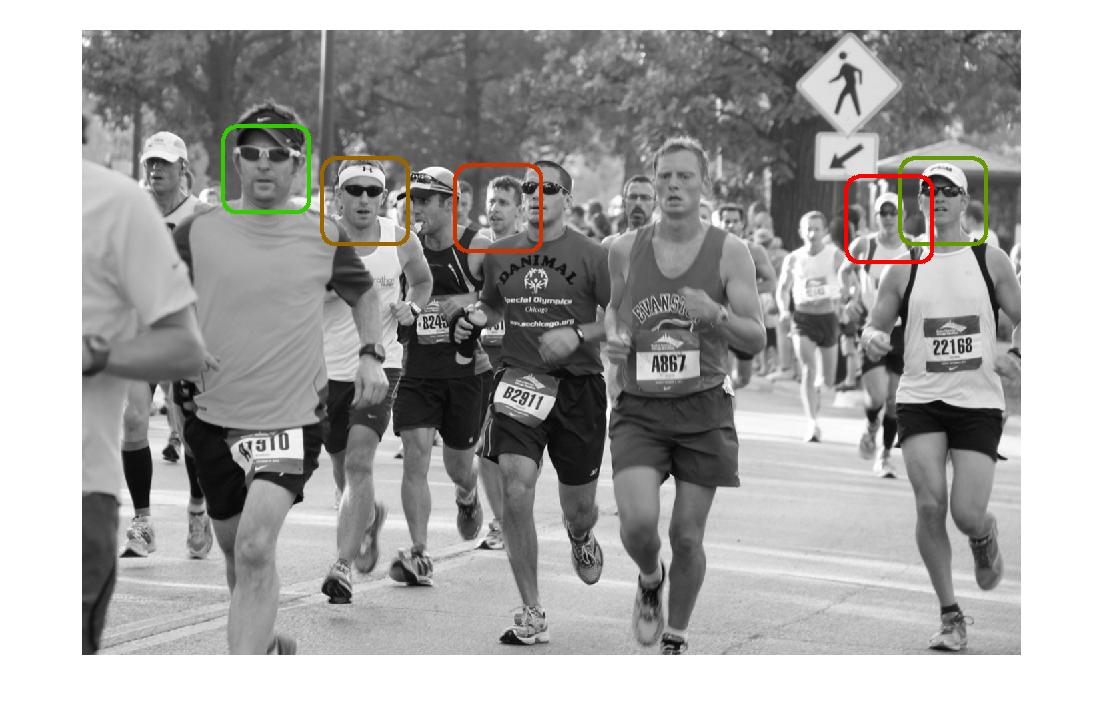}
\centering
\caption{\label{fig:frog}Detection at different scales}
\end{figure}
\begin{figure}
\centering
\includegraphics[width=1\textwidth]{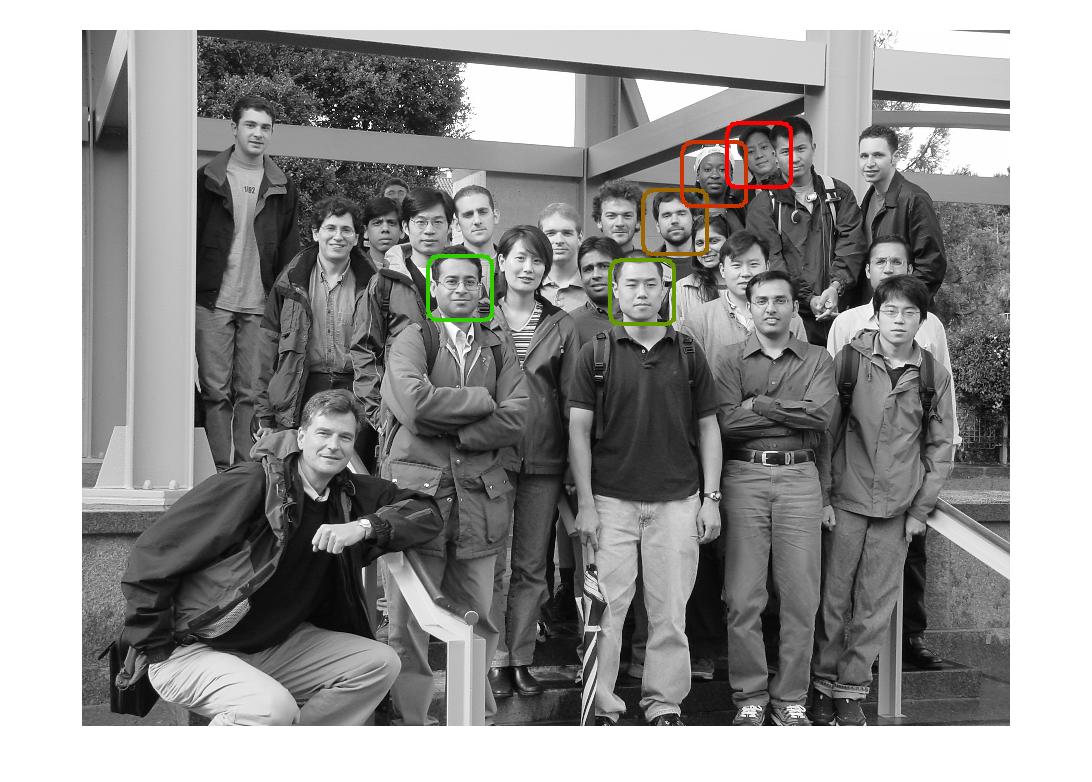}
\centering
\caption{\label{fig:frog}Detection at different scales}
\end{figure}
\end{enumerate}
\section{Reference}
\begin{enumerate}
\item Navneet Dalal and Bill Triggs, Histograms of Oriented Gradients for Human Detection, CVPR 2005
\item Carl Vondrick, Aditya Khosla, Tomasz Malisiewicz, Antonio Torralba, HOGgles: Visualizing Object Detection Features
\item Kai Briechle, Uwe D. Hanebeck, Template Matching using Fast normalized cross correlation.
\item Pedro Felzenszwalb,David McAllester,Deva Ramanan, A Discriminatively Trained, Multiscale, Deformable Part Model, IEEE Conference on Computer Vision and Pattern Recognition, 2008
\end{enumerate}
\end{document}